\documentclass[letterpaper, 10 pt, conference]{ieeeconf}  

\usepackage{amsmath}
\usepackage{amssymb} 

\renewcommand{\QED}{\QEDopen}

\usepackage[labelformat=simple]{subcaption}
\DeclareCaptionLabelSeparator{periodspace}{.\quad}
\usepackage{tikz}
\usetikzlibrary{arrows,matrix,positioning,patterns}
\usetikzlibrary{calc}
\usepackage{pgfplots} 
\usetikzlibrary{plotmarks}

\usepackage[utf8]{inputenc}
\usepackage{pgfplots}
\usepgfplotslibrary{groupplots}

\usepackage{url}
\usepackage{multirow}
\usepackage{colortbl}

\usepackage{algorithm} 
\usepackage{algpseudocode}

\DeclareSymbolFont{bbold}{U}{bbold}{m}{n}
\DeclareSymbolFontAlphabet{\mathbbold}{bbold}

\usepackage[labelformat=simple]{subcaption}
\DeclareCaptionLabelSeparator{periodspace}{.\quad}
\usepackage{listings}
\usepackage{xcolor}
\usepackage{cite}

\IEEEoverridecommandlockouts                              
\title{\LARGE \bf
Encoding Motion Primitives for Autonomous Vehicles using\\Virtual Velocity Constraints and Neural Network Scheduling}

\author{Mogens Graf Plessen
\thanks{{\tt\small \tt\small mgplessen@gmail.com}}
}

\renewcommand{\baselinestretch}{0.90}
\begin{document}

\maketitle
\thispagestyle{empty}
\pagestyle{empty}

%
%
%
%


\begin{abstract}
Within the context of trajectory planning for autonomous vehicles this paper proposes methods for efficient encoding of motion primitives in neural networks on top of model-based and gradient-free reinforcement learning. It is distinguished between 5 core aspects: system model, network architecture, training algorithm, training tasks selection and hardware/software implementation. For the system model, a kinematic (3-states-2-controls) and a dynamic (16-states-2-controls) vehicle model are compared. For the network architecture, 3 feedforward structures are compared including weighted skip connections. For the training algorithm, virtual velocity constraints and network scheduling are proposed. For the training tasks, different feature vector selections are discussed. For the implementation, aspects of gradient-free learning using 1 GPU and the handling of perturbation noise therefore are discussed. The effects of proposed methods are illustrated in experiments encoding up to 14625 motion primitives. The capabilities of tiny neural networks with as few as 10 scalar parameters when scheduled on vehicle velocity are emphasized.
\end{abstract}

\section{Introduction and Problem Formulation\label{sec_intro}}

\subsection{Motivation of a neural network approach for control}

Control methods that permit small sampling times, high-dimensional nonlinear nonconvex system models, and long planning horizons are very desirable. This also holds for autonomous driving. To underline complexity, for a sampling time of 0.01s and look-ahead time of 2s a time-based prediction horizon of 200 sampling instances is needed. This motivates \emph{offline} encoding of motion primitives in neural networks (NNs), before their employment online in combination with a reference waypoint selector. The main disadvantage is that typically high computational power is needed for the encoding of a large number of motion primitives in NNs.

\subsection{Problem formulation and contribution}

The problem addressed in this paper is to develop methods for efficient encoding of motion primitives in NNs. For clarity, it is stressed that the focus is here solely on aspects to \emph{improve the encoding process}, in particular, for dynamic vehicle models. Not discussed here are all recursive feasibility related problems that arise in closed-loop control, for scenarios unseen during training and model mismatches (subject of ongoing work). The following main contributions are made. First, specific virtual velocity constraints (VVC) are proposed. It is motivated how these must be handled differently for kinematic and dynamic vehicle models. For the latter, a specific 1-scalar network extension is suggested. Second, network scheduling is proposed, whereby vehicle velocity is used as scheduling variable. Third, various feature vector selections are discussed, before a preferred 4D choice is motivated. Fourth, 3 feedforward structures are compared including weighted skip connections. Fifth, details of the GPU-implementation and a full 16-states-2-controls dynamic vehicle model are discussed. Sixth, the benefits and capabilities of tiny NNs with as few as 10 parameters are illustrated.

\subsection{Related work}

Regarding the training method, this paper is based on the  TSHC-algorithm (task separation with hill climbing) \cite{plessen2017automating}, which itself was motivated by the ES-algorithm (evolution strategies) \cite{salimans2017evolution}. Note that ES can be considered a gradient-based algorithm since it performs stochastic gradient descent via an operation similar to a finite-difference approximation of the gradient \cite{such2017deep}, however, generating more robust policies~\cite{lehman2017more}. In contrast, TSHC is a truely gradient-free algorithm (hill climbing), which is designed specifically for deterministic encoding of motion primitives in NNs. This paper differs from \cite{plessen2017automating} according to above listed 6 contributions. 

Regarding control by motion primitives, this approach differs from methods derived from \cite{frazzoli1999hybrid}, which require online search of look-up tables, e.g., using a GPU for exhaustive search \cite{mcnaughton2011motion}. In contrast, when encoding motion primitives in a NN, explicit search is not required. Instead, a feature vector relating the current state to a next desired state (waypoint) is fed to the network to generate control commands. 

Regarding NN architectures, this paper extends recently proposed SCNs (structured control nets) \cite{srouji2018structured}, which combine a linear term mapping from network input to control channels additively with a nonlinear term resulting from a multilayer perceptron (MLP). In contrast, one of the discussed network architectures adds linearly weighted skip connections between all downstream layers. Skip connections are not new \cite{bishop1995neural}, but popular for learning very deep architectures \cite{he2016deep}.

Regarding vision-based end-to-end learning approaches \cite{pomerleau1989alvinn, bojarski2016end}, the proposed approach fundamentally differs in that it is founded on model-based training. This offers the advantage that certificates about learnt control performance can be provided by statement of (i) the vehicle-model used for training, and (ii) the encoded motion primitives (training tasks) and their associated low-dimensional feature vectors. In contrast, providing equivalent certificates for  vision-based end-to-end learning methods is in general much more difficult due to the high dimensionality of images. 

Finally, it is noted that a closed-loop control system based on encoded motion primitives must always be seen in combination with a reference setpoint selector that is determining waypoints or features to be fed to the NN, which must account for obstacles and thus primarily solve nonconvex optimization problems. For exemplatory approaches see \cite{kant1986toward, lozano2014constraint, srivastava2014combined, hu2018dynamic, liu2018elements, hoel2017evolutionary}. Explicit reference setpoint selection as well as a preceding perception module (fusing proprio- and exteroceptive sensor measurements) are not the focus of this paper. 

This paper is organized as follows. Vehicle models, network architectures, training algorithm and implementation details are discussed in Sections \ref{sec_veh_mdl}-\ref{sec_implemDetails}. Simulation experiments are reported in Section \ref{sec_expts}, before concluding.

\section{System Model\label{sec_veh_mdl}}

\subsection{Kinematic 3-states vehicle model\label{subsec_kin3statevehmdl}}

The equations of motion of a well-known simple kinematic vehicle model are: $\dot{x}=v \cos(\varphi)$, $\dot{y}=v \sin(\varphi)$ and $\dot{\varphi} = \frac{v}{L}\tan(\delta)$, with 
wheelbase $L$ (in simulations 2.69m). This model has 3 states (position-coordinates and heading) and 2 controls (steering angle $\delta$ and velocity $v$). Both controls are additionally constrained by absolute and rate actuation limits to emulate steering and 0-100/100-0km/h ac/deceleration performance of the dynamic vehicle model described next.

\subsection{Dynamic 16-states vehicle model\label{subsec_dyn16statevehmdl}}

Equations of motion of a 16-states dynamic vehicle model are derived by extending the bicycle model \cite{velenis2010steady}\footnote{Used as starting point since discussing a high-dimensional dynamic vehicle model and providing all hyperparameters for reproduction.} by aerodynamic friction forces, roll-, yaw- and four-wheel-dynamics:
 
{\allowdisplaybreaks
\vspace{-0.4cm}
\begin{subequations}
\begin{small}
\begin{align}
\dot{x} &= v_x \cos(\varphi) - v_y \sin(\varphi),\\
\dot{y} &= v_x \sin(\varphi) + v_y \cos(\varphi),\\
\dot{\varphi} &= \omega_\varphi,\\
\dot{v_x} &= \frac{1}{m}(\sum\nolimits_{j=1}^{4}F_{xj} - F_\text{xair}) + v_y\omega_\varphi,\\ 
\dot{v_y} &= \frac{1}{m}(\sum\nolimits_{j=1}^{4}F_{yj} - F_\text{yair}) - v_x\omega_\varphi,\\ 
\dot{\omega_\varphi} &= \frac{1}{I_z}(l_f(F_{y1}+F_{y2}) - l_r(F_{y3}+F_{y4}) +\dots\notag\\
 &  ~~~~~~~~l_w(F_{x2}+F_{x4}-F_{x1}-F_{x3})),\\ 
\dot{\psi} &= \omega_\psi,\\
\dot{\omega_\psi} &= \frac{1}{I_x}(l_w(F_{\eta 1}+F_{\eta 3}-F_{\eta 2}-F_{\eta 4}) + h\sum\nolimits_{j=1}^{4}F_{yj} ),\\ 
\dot{\phi} &= \omega_\phi,\\
\dot{\omega_\phi} &= \frac{1}{I_y}(l_r(F_{\eta 3}+F_{\eta 4}) - l_f(F_{\eta 1}+F_{\eta 2}) - h\sum\nolimits_{j=1}^{4}F_{xj} ),\\ 
\dot{\omega_1} &= \frac{1}{I_w}(T_{a1}-T_{b1} - r_eF_{xw1}),\\
\dot{\omega_2} &= \frac{1}{I_w}(T_{a2}-T_{b2} - r_eF_{xw2}),\\
\dot{\omega_3} &= \frac{1}{I_w}(-T_{b3} - r_eF_{xw3}),\\
\dot{\omega_4} &= \frac{1}{I_w}(-T_{b4} - r_eF_{xw4}),\\
\dot{\eta} &= v_\eta,\\
\dot{v_\eta} &= \frac{1}{m}(\sum\nolimits_{j=1}^4 F_{\eta j} - g).
\end{align}
\label{eqs_16dynvehmdl}
\end{small}
\end{subequations}}
Note that \eqref{eqs_16dynvehmdl} is based on the  Pacejka ``magic formula'' tyre model \cite{bakker1987tyre}. In addition, \cite{savaresi2010semi} and \cite{svendenius2007tire} were used for its derivation. Because of the general importance of \emph{models} for all model-based control and reinforcement learning algorithms, the entire \texttt{C++} code excerpt is provided in Appendix \ref{subsec_appendix_dyn16vehmdl}, including  all system parameters which are extended from \cite{velenis2010steady} and modified (among others) such that a 0-100/100-0km/h ac/deceleration performance of 7.4/3.8s is obtained. 

\emph{If-else} distinctions are convenient for model formulations and imply logical constraints. In the context of optimal control, these can be translated into integer linear inqualities \cite{williams2013model} yielding mixed-integer optimization problems. Note that logical constraints require no special treatment when encoding motion primitives in NNs via gradient-free learning.

Control commands are discussed. Suppose continuous control $a_t[i]\in[-1,1]$ for $i=0,1$ at sampling time $t$. Then, for the dynamic vehicle model,
\begin{subequations} 
\begin{align}
\delta_t &=\delta^\text{max}a_t[0],\label{eq_16dyn_delta}\\
T_{a,t} &=T_a^\text{min} + \frac{a_t[1]+1}{2}(T_a^\text{max}-T_a^\text{min}),\label{eq_16dyn_Ta}
\end{align}
\end{subequations}
before $T_{a,t}$ is distributed to drive- and brake-torques at the different wheels. In contrast, for the kinematic model only \eqref{eq_16dyn_delta} is used likewise, while \eqref{eq_16dyn_Ta} is replaced by $v_{t} =v^\text{min} + \frac{a_t[1]+1}{2}(v^\text{max}-v^\text{min})$, with $v^\text{max}$ and $v^\text{min}$ maximum and minimum velocity. For the dynamic model, both acceleration and deceleration are controlled via $a_t[1]$ (instead of, e.g., distinguishing front-wheel drive and 4 brake commands). This is done to compare kinematic and dynamic models with both having 2 controls. It implies that physical acceleration and braking actuators are never activated simultaneously. 

\subsection{Discussion of vehicle model and feature vector selection\label{subsec_discussVehMdlFeatureVec}}

Three more comments about vehicle models are made.  First, higher-fidelity vehicle models offer the potential of reducing control delays to a minimum. Therefore vehicles may be modeled to a degree such that the lowest-possible actuation commands are controlled, e.g., pulse-width modulated (PWM) signals. In contrast, simple kinematic models usually require additional cascaded low-level control for the mapping from PWM to velocity. For example, see \cite{plessen2017trajectory2} for spatial-based velocity control using a kinematic model.

Second, multiple system parameters which typically characterize higher-dimensional dynamic vehicle models offer means for robustifying control by encoding motion primitives for different system parameter settings.

Third, the choice of \emph{feature vector} $s_t$ input to the NN controller may not necessarily be affected by the vehicle model. Various feature vectors are considered. In this paper, for example, a 5-, 6- and 7D selection are considered. The first is defined as $s_t = [\frac{x^\text{goal}-x_t}{\Delta x_n},~\frac{y^\text{goal}-y_t}{\Delta y_n},~\frac{\varphi^\text{goal}-\varphi_t}{\Delta \varphi_n},~\frac{v_{x,t}}{v_{x,n}},~\frac{v_{x}^\text{goal}}{v_{x,n}}]$, relating states at time $t$ to desired goal pose. The latter two options (6D and 7D) add either only $a_{t-1}[0]$, or both $a_{t-1}[0]$ and $a_{t-1}[1]$, respectively. Normalization constants are employed, which throughout this paper are  selected as $[\Delta x_n,~\Delta y_n,~\Delta \varphi_n,~v_{x,n}]=[50,~3.5,~\frac{\pi}{2},~120/3.6]$ in SI-units. It is distinguished between $v_{x,t}$ and $v_{x}^\text{goal}$ to account for different velocity effects on the dynamics. Note that the first two options for $s_t$ are identical for both kinematic and dynamic vehicle model. However, the third option varies due to the different interpretation of $a_t[1]$ for the two models. The dimension of $s_t$ may influence the number of training tasks. This is since without a priori knowledge about meaningful training tasks, the simplest method to generate training tasks is to grid over the elements of $s_t$. 

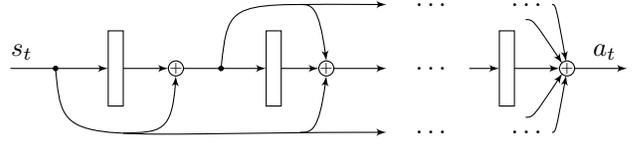
\begin{figure}
\centering
\vspace{0.3cm}
\begin{tikzpicture}
\node[color=black] (a) at (-1.15, 0.72) {$s_t$};
\draw[fill=black,solid] (-0.7,0.5) circle (0.03cm);
\draw[->,>=latex'] (-1.3, 0.5) -- (0, 0.5);
\draw[->,>=latex',color=black] (-0.7,0.5) .. controls (-0.6,-0.6) and (-0.6,-0.35) .. (3.7,-0.35);
\node[color=black] (a) at (4.32, -0.35) {$\dots$};
\draw[->,>=latex',color=black] (0.2,-0.36) .. controls (0.7,-0.35) and (0.85,-0.35) .. (0.9,0.4);
\draw[->,>=latex',color=black] (2.55,-0.35) .. controls (2.7,-0.35) and (2.8,-0.35) .. (2.9,0.4);
\draw[fill=white,solid] (0,0) rectangle (0.2,1);
\draw[->,>=latex'] (0.2, 0.5) -- (0.8, 0.5);
\draw (0.9,0.5) circle (0.1cm);
\node[color=black] (a) at (0.9, 0.5) {\tiny{$+$}};
\draw[->,>=latex'] (1, 0.5) -- (2.1, 0.5);
\draw[fill=black,solid] (1.5,0.5) circle (0.03cm);
\draw[->,>=latex',color=black] (1.5,0.5) .. controls (1.6,1.45) and (1.6,1.35) .. (3.7,1.35);
\node[color=black] (a) at (4.32, 1.35) {$\dots$};
\draw[->,>=latex',color=black] (2.55,1.35) .. controls (2.7,1.35) and (2.8,1.35) .. (2.9,0.6);
\draw[fill=white,solid] (2.1,0) rectangle (2.3,1);
\draw[->,>=latex'] (2.3, 0.5) -- (2.8, 0.5);
\draw (2.9,0.5) circle (0.1cm);
\node[color=black] (a) at (2.9, 0.5) {\tiny{$+$}};
\draw[->,>=latex'] (3, 0.5) -- (3.7, 0.5);
\node[color=black] (a) at (4.32, 0.5) {$\dots$};
\draw[->,>=latex',color=black]  (5.9,-0.35) .. controls (5.95,-0.35) and (5.95,-0.35) .. (6.088,0.409);
\node[color=black] (a) at (5.59, -0.35) {$\dots$};
\draw[->,>=latex',color=black]  (5.55,-0.15) .. controls (5.65,-0.15) and (5.65,-0.15) .. (6.038,0.435);
\draw[->,>=latex',color=black]  (5.9,1.35) .. controls (5.95,1.35) and (5.95,1.35) .. (6.088,0.591);
\node[color=black] (a) at (5.59, 1.35) {$\dots$};
\draw[->,>=latex',color=black]  (5.55,1.15) .. controls (5.65,1.15) and (5.65,1.15) .. (6.038,0.565);
\draw[->,>=latex'] (4.8, 0.5) -- (5.2, 0.5);
\draw[fill=white,solid] (5.2,0) rectangle (5.4,1);
\draw[->,>=latex'] (5.4, 0.5) -- (6.0, 0.5);
\draw (6.1,0.5) circle (0.1cm);
\node[color=black] (a) at (6.1, 0.5) {\tiny{$+$}};
\draw[->,>=latex'] (6.2, 0.5) -- (6.9, 0.5);
\node[color=black] (a) at (6.6, 0.72) {$a_t$};
\end{tikzpicture}
\caption{The concept of FSCNs is sketched for visualization. For the general description, see \eqref{eq_FSCN} and note the superposition of nonlinear and linear terms.}
\label{fig_FSCN}
\vspace{-0.5cm}
\end{figure}

\section{Neural Network Architecture\label{sec_netw_architecture}}

The processing of feature vector $s_t$ by NNs is discussed.

\subsection{Fully Structured Control Nets (FSCNs)}

\emph{Fully structured control nets} (FSCNs) are introduced as
\begin{subequations}
\begin{align}
s_\text{in}^{(0)} &= s_t,\\
s_\text{in}^{(l)} &= s_\text{out}^{(l-1)} + \sum_{j=0}^{l-1}s_\text{in}^{(j)}K^{(jl)},~ \forall l=1,\dots,N_l-1,\label{eq_FSCN_sinl}\\
s_\text{out}^{(l)} &= \text{tanh}\left(s_\text{in}^{(l)} W^{(l)} + b^{(l)}\right),~\forall l=0,\dots,N_l-1,\\
a_t &= s_\text{out}^{(N_l-1)} + \sum_{j=0}^{N_l-1}s_\text{in}^{(j)}K^{(jN_l)} + c^{(N_l)},\label{eq_FSCN_at}
\end{align}
\label{eq_FSCN}
\end{subequations}
with parameters to learn $W^{(l)}\in\mathbb{R}^{N^{(l)}\times N^{(l+1)}}$, $b^{(l)}\in\mathbb{R}^{1\times N^{(l+1)}}$, $K^{(jl)}\in\mathbb{R}^{N^{(j)}\times N^{(l+1)}}$ and $c^{(N_l)}\in\mathbb{R}^{1\times n_a}$, $\forall l=0,\dots,N_l-1$ and $j=0,\dots,N_l-1$ with $j<l$. For the remainder of this paper all parameters to be learnt shall be summarized in a vector denoted by $\theta$ initialized by small zero-mean Gaussian noise (with a standard deviation of 0.001). For illustration of the concept of FSCNs see also Fig. \ref{fig_FSCN}. FSCN design choices are the number of layers $N_l$ and the number of units per layer $N^{(l)},~\forall l=1,\dots,N_l-1$. Note that $N^{(0)}=n_s$ and $N^{(N_l)}=n_a$ are fixed as the dimensions of feature vector $s_t$ and controls $a_t$, respectively.

\subsection{Discussion of neural network architectures}

Five remarks are made. First, FSCNs \eqref{eq_FSCN} extend recently proposed \emph{structured control nets} (SCNs) \cite{srouji2018structured}, which for comparison read: $s_\text{in}^{(0)} = s_t$, $s_\text{out}^{(l)} = \text{tanh}(s_\text{in}^{(l)} W^{(l)} + b^{(l)}),~\forall l=0,\dots,N_l-1,$ and $a_t = s_\text{out}^{(N_l-1)} + s_\text{in}^{(0)}K^{(0N_l)} + c^{(N_l)}$. Thus, in contrast to \eqref{eq_FSCN}, SCNs only add one linear term from network input to output. In Sect. \ref{subsec_expt3}, both architectures plus the standard \emph{multilayer perceptron} (MLP) are compared. MLPs itself are identical to SCNs minus the additional linear term.

Second, weighted skip connections are introduced as in \eqref{eq_FSCN_sinl} and \eqref{eq_FSCN_at}. Since these weights are initialized by small Gaussian noise with zero-mean, FSCNs initially resemble MLPs. Alternatively, initialization as identity mappings (plus small Gaussian noise) rather than around zero were also tested (then more emphasizing the skip-aspect) but not found to accelerate learning at the early training phase. 

Third, in this paper small NNs (with few parameters) that still enable encoding of all motion primitives (training tasks) are desired. Small NNs are preferrable since they (i) permit faster network evaluation, which is favorable for both faster offline training and online control execution, and (ii) reduce large hardware storage requirements for parameters. For perspective, in \cite{such2017deep} huge 4M+ parameter networks are mentioned for playing Atari games (requiring image processing). In contrast, we here seek to reduce the number of parameters as much as possible. The benefits of small networks become most apparent when training with limited computational ressources. The effects on training times are demonstrated in the experiments of Sect. \ref{subsec_expt2}.

Fourth, in contrast to MLPs, in general $a_t\in [-1,1]$ in \eqref{eq_FSCN_at} cannot be guaranteed. This is because of its affine term. In experiments capping and an additional tanh$(\cdot)$ activation were tested but found to not accelerate learning (on the contrary). Note that $a_t\in [-1,1]$ is ultimately ensured through physical actuator absolute and rate constraints.

Fifth, as will be shown, for the encoding of motion primitives based on  \emph{dynamic} vehicle models that do not control velocity directly, a NN-extension was found to be very useful for the handling of spatial virtual velocity constraints (VVCs). These constraints and the corresponding network extension are presented next and consequently applied to all 3 NNs discussed: FSCNs, SCNs and MLPs.

\section{Training Algorithm \label{sec_trainAlg}}

This section states key aspects of the proposed algorithm for efficient encoding of motion primitives in above NNs.

\subsection{Virtual Velocity Constraints and Network Extension\label{subsec_VVCs}}

The notion of \emph{virtual velocity constraints} (VVCs) is adopted from \cite[Sect. III-B]{plessen2017automating}. However, due to the dynamic vehicle model used here and to remove 1 hyperparameter, modifications on VVCs are made as follows. First, let 
\begin{equation}
\tilde{v}_t = v^\text{min} + \frac{a_t[1]+1}{2}(v^\text{max}-v^\text{min}),
\end{equation}
with $a_t[1]$ being output of a NN for both cases of training on a kinematic and dynamic vehicle model, and let
\begin{equation}
v_\text{VVC}^\text{max} = v_x^\text{goal} + \frac{5}{3.6},~\quad\text{and}\quad~v_\text{VVC}^\text{min} = v_x^\text{goal} - \frac{5}{3.6}.\label{eq_def_vVCmaxmin}
\end{equation}
Second, if $\tilde{v_t}>v_\text{VVC}^\text{max}$ or $\tilde{v_t}<v_\text{VVC}^\text{min}$, project $\tilde{v_t}=v_\text{VVC}^\text{max}$ and $\tilde{v_t}=v_\text{VVC}^\text{min}$, respectively. Third, and now differentiating between the cases of training on a kinematic and dynamic vehicle model, set
\begin{equation}
a_t[1] = \frac{\tilde{v}_t - v^\text{min}}{v^\text{max}-v^\text{min}}2 - 1,
\end{equation}
and 
\begin{equation}
a_t[1] = a_v^\text{thres} + \text{tanh}\left(   
\theta_\text{VVC}(v_{x,t}-\tilde{v}_t)
\right),\label{eq_VCs_at1_dyn}
\end{equation}
for the former and latter cases, respectively. Here,  $a_v^\text{thres}=-1-2\frac{T_a^\text{min}}{T_a^\text{max}- T_a^\text{min}}$ is such that $T_{a,t}=0$ and $\theta_\text{VVC}$ is a scalar parameter to be learnt. However, only when training based on a dynamic vehicle model. Ultimately, $a_t[i]$ for $i=0,1$ are applied to physical actuator absolute and rate constraints accounting for $a_{t-1}[i]$ at the previous sampling time.

Several comments are made. First, the VVCs of \eqref{eq_def_vVCmaxmin} are spatially independent of goal proximity. This has several benefits: (i) due to the $\pm\frac{5}{3.6}$-margin feasibility can be guaranteed even for training tasks that demand, e.g., only a small lateral displacement of the vehicle for a desired starting and end velocity of 0km/h, and (ii) no hyperparameters and additional measures are required to threshold spatial goal proximity.

Second, the $\pm\frac{5}{3.6}$-margin around $v_x^\text{goal}$ is a heuristic choice. In general, it may be regared as a hyperparameter. However, here it is considered as fixed and to be interpreted as tolerable limitedly small velocity variation (for over/undershoots). The velocity corridor provided by $[v_\text{VVC}^\text{min},~v_\text{VVC}^\text{max}]$ encourages to always quickly and monotonously approach the target velocity. This property is (i) in general desirable, especially for throughput maximization and when it is encouraged and permitted by traffic to drive at speed limits (e.g., for urban driving), and (ii) encourages at most one velocity-sign change for the reaching of the goal velocity and state.

Third, VVCs can be regarded as a filter for the $a_{t}[1]$-output from a NN. As indicated in \eqref{eq_VCs_at1_dyn}, for training on the dynamic vehicle model scalar $\theta_\text{VVC}$ must be learnt in addition to the other NN parameters. Since velocity is not controlled directly the encoding of motion primitives based on dynamic vehicle models is more complicated. In preliminary tests a variety of alternative filter functions were tested. The simple form of \eqref{eq_VCs_at1_dyn} was found to be suitable. It requires just \emph{one} scalar parameter and resembles a P-controller with nonlinear activation function. Note that $a_{t}[1]=a_v^\text{thres}$ when $v_{x,t}=\tilde{v}_t$.

Fourth and to summarize, VVCs are motivated (i) to accelerate learning, and (ii) to avoid velocity-over/undershoots until reaching of desired goal-poses. The benefits of VVCs for both training on a kinematic and a dynamic vehicle model are illustrated in the experiments of Sect. \ref{subsec_expt1}. They are found to be essential for efficient encoding of motion primitives in NNs, especially when training on sparse rewards as motivated in \cite[Sect. III-B]{plessen2017automating}.

\subsection{Task Separation with Hill Climbing\label{subsec_TSHC}}

The gradient-free TSHC algorithm with refinement step \cite{plessen2017automating} is used for training, whereby perturbation hyperparameter $\sigma_\text{pert}\sim \mathcal{U}[10,1000]$ is selected randomly according to a uniform distribution at every parameter iteration to reduce the number of difficult to select hyperparameters (which would occur when instead using a fixed or adaptive $\sigma_\text{pert}$). This is relevant for the generation of $n$ parameter solution candidates
\begin{equation}
\theta_i\leftarrow \theta + \sigma_\text{pert}\xi_i, \label{eq_def_thetai}
\end{equation}
with Gaussian distributed $\xi_i\sim\mathcal{N}(0,I)$ for all $i=1,\dots,n$.

To summarize, all hyperparameters remaining for the training algorithm are: $N_\text{restarts}$, $N_\text{iter}^\text{max}$, $n$, $T^\text{max}$ and $\epsilon$. The number of restarts $N_\text{restarts}$ and maximum number of iterations per restart $N_\text{iter}^\text{max}$ may be selected according to desired total training time. When training on a GPU we may select $n$ as the product of the number of blocks and threads per block used for asynchronous training. In general a feasibility guarantee of all motion primitives can be given since training is conducted obstacle-free, which encourages to select the maximum number of permitted time-steps to solve a training task $T^\text{max}$ large. On the other hand, an unnecessarily conservative choice prolongs training. Tolerances $\epsilon=[\epsilon_d,~ \epsilon_\varphi,~\epsilon_v]$ indicate when a specific training task goal-pose is reached. 

Above comments underline a benefit of training by TSHC. That is its simplicity and interpretability of hyperparameters. Assuming large computational power being available, (i) all of $N_\text{restarts}$, $N_\text{iter}^\text{max}$, $n$ and $T^\text{max}$ should be large, and (ii) $\epsilon$ should be small. Then, the only tunable hyperparameter remaining is $\sigma_\text{pert}$. In practice, it was found that it should be selected sufficiently large to enable enough exploration in the NN parameter space. Hence, the choice $\sigma_\text{pert}\sim \mathcal{U}[10,1000]$.

\subsection{Neural Network Scheduling based on Vehicle Velocity\label{subsec_netwScheduling}}

The TSHC algorithm used to encode a \emph{set} of motion primitives in a NN was discussed above. Now it is proposed to partition a large set of motion primitives into \emph{subsets} of motion primitives \emph{scheduled on vehicle velocity}. Then, NNs may be learnt separately for each of these subsets by separate applications of the TSHC algorithm. As further demonstrated in the experiments of Sect. \ref{subsec_expt4}, this offers the advantage that learning effort can be adapted to difficulty of corresponding subsets of training tasks, e.g., using different network parametrizations and hyperparameters. Consequently, the overall time to learn the entire set of motion primitives can be reduced significantly. A disadvantage is a natural increase in the total number of network parameters. However, the former advantages clearly outweigh the latter disadvantage. This is since, as will be shown in the experiments of Sect. \ref{sec_expts}, tiny NNs can be used to encode many motion primitives.

\section{Implementation details\label{sec_implemDetails}}

All methods are implemented in \texttt{Cuda C++}. Training is conducted on 1 GPU. Three more comments are made. First, a self-imposed guideline was to implement \emph{library-free} code for the NN controller (such that in principle it could then run library-free on embedded hardware). Therefore, the tanh$(\cdot)$-function is approximated by an implementation of an Lambert's continued fraction series expansion. 

Second, as outlined in Sect. \ref{subsec_TSHC}, parameter candidates $\theta_i$ are generated by affine perturbations with zero-mean Gaussian noise and spherical variance $\sigma_\text{pert}^2$. Therefore, uniform random numbers are first generated according to \cite{park1988random}, before Gaussian random variables are generated based on the Box-Muller method \cite{box1958note}. One instance of the latter method simultaneously generates two scalar Gaussian variables. Both are used to generate consecutive entries in $\theta_i$. This enables library-free code. Furthermore, the same methods for uniform and Gaussian random variables are used on \emph{both} GPU and CPU host. Thus, only $\theta$ from \eqref{eq_def_thetai} and scalar random seeds need to be passed to the $n$ GPU kernels (workers), before parameter candidates are then generated directly on the GPU. A ranking of the performance of these $n$ workers and knowledge of their seed numbers then permits to reconstruct the best $i^\star$ and corresponding $\theta_{i^\star}$ on the CPU host.  

Third, for final experiments each of $n$ GPU kernels implements one of $n$ workers such that $n$ parameter candidates are tested in parallel. For completeness, nested parallelization with solving training tasks in parallel for each parameter candidate is in general also possible. This was also tested and implemented by using Cuda's \texttt{atomicAdd}$(\cdot)$-function and an algebraic mapping to reconstruct a specific training task from a kernel's thread index. For our scenario with 1 GPU this method did, however, not accelerate training. In contrast, the preferred method testing $n$ parameter candidates in parallel implies that all $N_\text{tasks}$ training tasks are tested for each parameter candidate and GPU kernel, before a cumulative score is returned to the CPU host. Training tasks are generated directly on the GPU within nested for-loops instead of precomputation to minimize memory requirements.

\section{Simulation Experiments\label{sec_expts}}

Tolerances indicating the reaching of a desired goal pose are set as $\epsilon_d=0.25$m and $\epsilon_v=5$km/h. As will be discussed, $\epsilon_\varphi$ is relevant only for Experiment 1 and is set as $5^\circ$.

\subsection{Experiment 1: Effects of $s_t$, vehicle model and VVCs\label{subsec_expt1}}

\begin{table}
\vspace{0.3cm}
\centering
\caption{Summary of Experiment 1, see Section \ref{subsec_expt1} for discussion. Results for the 16-states-2-controls dynamic and 3-states-2-controls kinematic vehicle model are differentiated by 16$\times$2 and 3$\times$2, respectively. \label{tab_ex1}}
 \def\arraystretch{0.8}
 \begin{tabular}{|p{0.005cm}|l|c|c|c|c|c|}
 \hline
& & $N_\text{tasks}^\star$ & $P^\star$ & $N_\text{param}^\text{netw}$ & $T_\text{learn}^\text{tot}$ & $N_\text{rest}^\star$ \\
\hline\hline 
\multirow{6}{*}{\hspace{-0.1cm}\rotatebox[origin=c]{90}{16$\times$2}} 
& (s5, VVCs) & 79 & - & 30 & 320.9 & 0\\
& (s5,-) & 68 & - & 29 & 341.9 & 0 \\\cline{2-7}
& \textbf{(s6, VVCs)} & \textbf{125} & \textbf{-1959.3} & \textbf{34} & \textbf{379.1} & \textbf{10}\\
& (s6,-) & 79 & - & 33 & 330.0 & 0 \\\cline{2-7}
& (s7, VVCs) & 125 & -1959.3 & 38 & 342.5 & 9 \\
& (s7,-) & 100 & - & 37 & 330.5 & 0 \\ 
\hline\hline  
\multirow{4}{*}{\hspace{-0.1cm}\rotatebox[origin=c]{90}{3$\times$2}} 
& (s5, VVCs) & 118 & - & 29 & 51.3 & 0 \\
& (s5,-) & 77 & - & 29 & 51.9 & 0 \\\cline{2-7}
& \textbf{(s6, VVCs)} & \textbf{125} & \textbf{-1956.3} & \textbf{33} & \textbf{52.4} & \textbf{10} \\
& (s6,-) & 88 & - & 33 & 54.8 & 0 \\
\hline 
\end{tabular}
\vspace{-0.3cm}
\end{table}

Experiment 1 is characterized by (i) a comparison of the 5-, 6- and 7D feature vectors from Section \ref{subsec_discussVehMdlFeatureVec} here abbreviated as s5, s6 and s7, (ii) $T^\text{max}=500$, (iii) FSCNs with 1 hidden layer and 1 hidden unit, i.e., [5,1,2], [6,1,2] and [7,1,2] (with each number in brackets indicating the number of units per layer), (iv) $(N_\text{restarts},N_\text{iter}^\text{max})=(10,20)$, (v) solving all tasks at once (i.e., without network scheduling), and (vi) training tasks according to $v_{x,0}\in\{0,5,\dots,120\}$~[km/h], $v_x^\text{goal}=v_{x,0}+\{-25,-12.5,0,12.5,25\}$ (capped between 0 and 120km/h), $x^\text{goal}=v_{x,0}t_\text{task} + \frac{1}{2}0.6a^\text{max}t_\text{task}^2$ with $t_\text{task}=\frac{v_x^\text{goal}-v_{x,0}}{0.8a^\text{max}}$ and $a^\text{max}=\frac{100}{3.6\cdot 7.4}$ if $v_x^\text{goal}\geq v_{x,0}$ as well as $x^\text{goal}=v_{x,0}t_\text{task} + \frac{1}{2}0.6a^\text{min}t_\text{task}^2$ with $t_\text{task}=\frac{v_x^\text{goal}-v_{x,0}}{0.8a^\text{min}}$ and $a^\text{min}=-\frac{100}{3.6\cdot 3.8}$ if $v_x^\text{goal}< v_{x,0}$, $y^\text{goal}=0$, $\varphi^\text{goal}=0$, $a_{-1}[0]=0$ and $a_{-1}[1]=a_v^\text{thres}$. This yields a total number of $N_\text{tasks}=125$ training tasks, which are selected to analyze longitudinal control (to focus on VVCs effects) and to ensure a maximum look-ahead time of less than 2.5s. Thus, for the selected $T^\text{max}=500$ (in combination with a sampling time of 0.01s) all training tasks are guaranteed to be learnable.  

\newlength\figureheight
\newlength\figurewidth 
\setlength\figureheight{6.5cm}
\setlength\figurewidth{6.5cm}
\begin{figure}
\vspace{0.3cm}
\centering%
\includegraphics[scale=1.0]{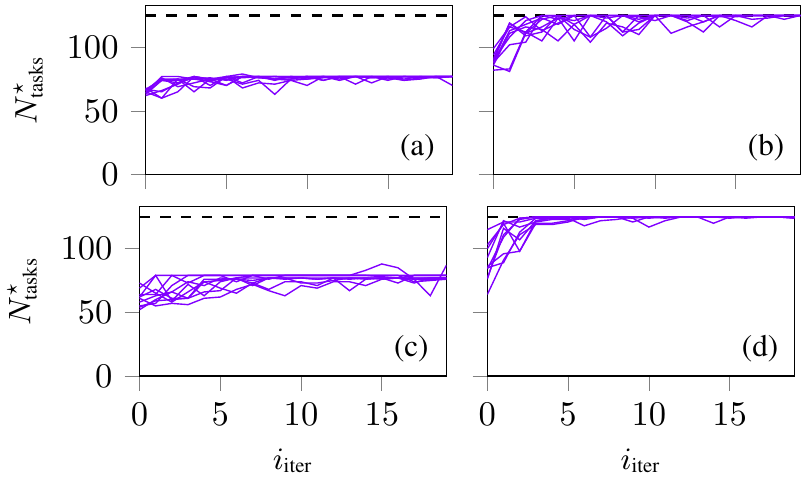}
\caption{Experiment 1. The benefits of VVCs are demonstrated for both dynamic (a-b) and kinematic vehicle models (c-d). The learning results \emph{without} VVCs are shown in (a) and (c). The black-dashed line indicates the desired number of motion primitives to be learnt. For each subplot, $N_\text{tasks}^\star(i_\text{iter}),\forall i_\text{iter}=0,\dots,N_\text{iter}^\text{max}$ is displayed for all restarts $N_\text{restarts}=10$. }
\label{fig:ex1_4subplts}
\vspace{-0.5cm}
\end{figure}

\setlength\figureheight{6.5cm}
\setlength\figurewidth{6.5cm}
\begin{figure}
\vspace{0.3cm}
\centering%
\includegraphics[scale=1.0]{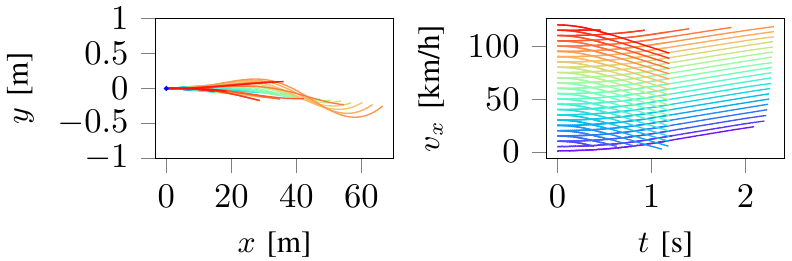}
\caption{Experiment 1. Visualization of learnt performance (trajectories and velocity profiles) for the dynamic vehicle model and (s6, VVCs)-setting. }
\label{fig:ex1_2subplts}
\vspace{-0.4cm}
\end{figure}

Results are summarized in Table \ref{tab_ex1}, Fig. \ref{fig:ex1_4subplts} and \ref{fig:ex1_2subplts}, whereby $N_\text{tasks}^\star$ denotes the number of tasks solved to $\epsilon$-precision, $P^\star$ [m] the negative (hill climbing convention) accumulated pathlength, $N_\text{param}^\text{netw}$ the number of network parameters (with $\theta_\text{VVC}$), $T_\text{learn}^\text{tot}$~[s] the total learning time and $N_\text{rest}^\star$ the number of restarts (out of $N_\text{restarts}$) for which all tasks could be solved.

Several observations can be made. First, VVCs clearly improve learning progress for both kinematic and dynamic vehicle models, see Fig. \ref{fig:ex1_4subplts}. Second, the inclusion of $a_t[0]$ in feature vector $s_t$ clearly helps: compare s6 and s7 (solving all tasks) vs. s5 (omitting $a_t[0]$ from $s_t$ and not solving all tasks). Note also the robustness for the former cases w.r.t. restarts. As $N_\text{rest}^\star=10$ indicates, for s6 all 125 tasks are solved for all restarts. Third, despite varying $\sigma_\text{pert}$ randomly according to Section \ref{subsec_TSHC}, an evolution of parameters over $i_\text{iter}$ and corresponding learning progress can be observed, see Fig. \ref{fig:ex1_4subplts} (b) and (d). Fourth, significantly faster training times are observed for the kinematic in comparison to the dynamic vehicle model. This is since model simulations are much more complex for the latter, which accumulates to longer $T_\text{learn}^\text{tot}$. Fourth, as Fig. \ref{fig:ex1_2subplts} shows within limited $T_\text{learn}^\text{tot}$ desirable steering trajectories with small maximum lateral overshoot of only 0.41m (among all tasks) from optimal $y=0$  are learnt. Fig. \ref{fig:ex1_2subplts} further indicates that monotonous velocity profiles are learnt. Finally, note that FSCNs with only one hidden layer and unit were sufficient to encode all training tasks.

\subsection{Experiment 2: Effects of network size for FSCNs\label{subsec_expt2}}

Experiment 2 is characterized by (i) feature vector
\begin{equation}
s_t = \begin{bmatrix} \frac{y^\text{goal}-y_t}{\Delta y_n} & \frac{v_{x,t}}{v_{x,n}} & \frac{v_{x}^\text{goal}}{v_{x,n}} & a_{t-1}[0] \end{bmatrix},\label{eq_def_st_4D} 
\end{equation}
and (ii) $T^\text{max}=500$, (iii) a comparison of FSCNs for different numbers of hidden layers and units per hidden layer, (iv) $(N_\text{restarts},N_\text{iter}^\text{max})=(10,20)$, (v) attempting to solve a total of $N_\text{tasks}=585$ training tasks at once (i.e., without network scheduling), which are generated by gridding $y^\text{goal}\in[0,3.5]$ and $v_{x,0}\in[0,130]$ with uniform spacings of 0.25m and 10km/h, respectively, $v_{x}^\text{goal}=v_{x,0}+\{-10,0,10\}$ for all $v_{x,0}$,  $a_{-1}[0]=0$ and $a_{-1}[1]=a_{v}^\text{thres}$. 
\begin{table}
\vspace{0.3cm}
\centering
\caption{Experiment 2. The number of hidden layers and number of units per hidden layer (uniform among multiple hidden layers) is indicated by rows and columns,  respectively. Throughout, a 4D feature vector $s_t$ is mapped to 2 controls. Thus, to give an example, row 2 and column labeled 8 indicate a FSCN-[4,8,8,2] network. For each row$\times$column combination $N_\text{tasks}^\star/N_\text{param}^\text{netw}/T_\text{learn}^\text{tot}$ are reported. For each row the setting with maximum $N_\text{tasks}^\star$ is in bold. Note how quickly $T_\text{learn}^\text{tot}$~[s] rises along the column-dimension with increasing number of units per hidden layer. \label{tabGrid_ex2}}
 \def\arraystretch{0.8}
 \begin{tabular}{|l|c|c|c|c|}
 \hline
  & 1 & 2 & 4 & 8 \\
\hline\hline 
1 & 494/26/3674 & 531/39/4442 & \textbf{550/65/4744} & 548/117/6278 \\ 
2 & 445/35/3687 & 536/61/4476 & 533/125/7073 & \textbf{554/301/10026} \\
3 & 479/45/4215 & 496/87/5780 & \textbf{538/201/8030} & 535/549/13159\\ 
\hline
\end{tabular}
\vspace{-0.cm}
\end{table}

Results are summarized in Table \ref{tabGrid_ex2}. Several comments are made. First, note how $T_\text{learn}^\text{tot}$ quickly rises with increasing network size. Experiment 2 was setup such that neither of the solutions in Table \ref{tabGrid_ex2} solved all 585 tasks to (i) better illustrate the effects of different network sizes on FSCN-performance, and to (ii) underline the role of $T^\text{max}$ in combination with Experiment 3, which treats the exact same training tasks. 

Second, \eqref{eq_def_st_4D} differs from the 5-, 6- and 7D feature vectors discussed in Sect. \ref{subsec_discussVehMdlFeatureVec} and Experiment 1. Feature vector $s_t$ according to \eqref{eq_def_st_4D} is our preferred choice. This is motivated for three reasons. (i) While experimenting with different training task generation schemes it was found that including $x^\text{goal}$- and $\varphi^\text{goal}$-related components in $s_t$ made training task setup much more complicated for control tasks requiring lateral motion. Gridding that guarantees feasibility, simultaneously enables to limit learning effort (limitedly small $T^\text{max}$), generalizes enough and therefore does not require much manual tuning is difficult. In contrast, gridding over the components of \eqref{eq_def_st_4D} is relatively straightforward. (ii) Since the NN controller is ultimately envisioned in combination with obstacle avoidance feasibility checks along forward simulated trajectories in a receding horizon fashion, focusing on $y^\text{goal}$-related training tasks appears suitable therefore and sufficient to encode lateral motion agility in NNs. This discussion is subject of ongoing work, see also Experiment 4 and Fig. \ref{fig:ex4_traj} for illustration and Sect. \ref{sec_conclusion}. (iii) Furthermore, focus on $y$ enables \emph{control mirroring} w.r.t. steering. This permits to limit training tasks to $y^\text{goal}\geq 0$ and thus use free training capacities to increase, e.g., lateral spacing resolution.

\subsection{Experiment 3: Aspects of different network architectures\label{subsec_expt3}}

\begin{table}
\centering
\caption{Experiment 3. The solutions for FSCN and MLP that solve all 585 tasks and yield the largest $P^\star$ are emphasized in bold, respectively.  \label{tab_ex3}}
\vspace{-0.3cm}
\begin{align*}
\def\arraystretch{0.8}
& \begin{tabular}{|l|c|c|c|c|c|}
\hline
FSCN & $N_\text{tasks}^\star$ & $P^\star$ & $N_\text{param}^\text{netw}$ & $T_\text{learn}^\text{tot}$ & $N_\text{rest}^\star$\\
\hline\hline 
$[4,1,2]$ & \textbf{585} & \textbf{-12814} & \textbf{26} & \textbf{9526} & \textbf{1}  \\ 
$[4,2,2]$ & 585 & -14644 & 39 & 13191 & 5 \\
$[4,4,2]$ & 585 & -13470 & 65 & 15803 & 7 \\ 
\hline
\end{tabular}
\\
\def\arraystretch{0.8}
& \begin{tabular}{|l|c|c|c|c|c|}
\hline
SCN~\cite{srouji2018structured} & $N_\text{tasks}^\star$ & $P^\star$ & $N_\text{param}^\text{netw}$ & $T_\text{learn}^\text{tot}$ & $N_\text{rest}^\star$\\
\hline\hline 
$[4,1,2]$ & 577 & -23665 & 20 & 3685 & 0\\
$[4,2,2]$ & 560 & -15908 & 27 & 3756 & 0\\
$[4,4,2]$ & 580 & -22084 & 41 & 3764 & 0\\
\hline
\end{tabular}
\\
\def\arraystretch{0.8}
& \begin{tabular}{|l|c|c|c|c|c|}
\hline
MLP & $N_\text{tasks}^\star$ & $P^\star$ & $N_\text{param}^\text{netw}$  & $T_\text{learn}^\text{tot}$ & $N_\text{rest}^\star$\\
\hline\hline 
$[4,1,2]$ & 577 & -13535 & 10 & 3554 & 0\\
$[4,2,2]$ & 566 & -14273 & 17 & 3602 & 0\\
$[4,4,2]$ & \textbf{585} & \textbf{-11562} & \textbf{31} & \textbf{3725} & \textbf{2}\\
\hline
\end{tabular}
\end{align*}
\vspace{-0.3cm}
\end{table}

Training setup is identical to Experiment 2 except that $T^\text{max}$ is increased from 500 to 1000. Different network architectures (FSCNs, SCNs~\cite{srouji2018structured} and MLPs) are compared for 3 different network sizes ([4,1,2], [4,2,2] and [4,4,2]). Results are summarized in Table \ref{tab_ex3}. The following observations are made. First, increasing $T^\text{max}$ enables to solve all tasks. While for Experiment 2 and $T^\text{max}=500$ even the largest FSCN could not solve all 585 tasks, for Experiment 3 and $T^\text{max}=1000$ every and even the smallest FSCN can solve all tasks. In contrast, none of the SCNs and only MLP-[4,4,2] could solve all $N_\text{tasks}=585$. Note that the best $P^\star$-result is obtained for MLP-[4,4,2].

Second, significantly larger $T_\text{learn}^\text{tot}$ are observed for FSCNs in comparison to SCNs and MLPs. One reason is that for, e.g., FSCN-[4,1,2] 178 out of $N_\text{restarts} N_\text{iter}^\text{max}=200$ best parameter settings solved less than 500 tasks. In contrast, for MLP-[4,1,2] only 68 out of $200$ solved less than 500 tasks (despite none solving all 585). This was a recurring observation. Thus, on average per GPU-call MLPs solved more tasks, which reduced overall $T_\text{learn}^\text{tot}$. 

Third, for the same number of hidden layers and units per hidden layer, MLPs always have fewer parameters $N_\text{param}^\text{netw}$ than SCNs, which themselves have fewer than FSCNs.
 
To summarize, based on above observations, in particular on $T_\text{learn}^\text{tot}$ and $N_\text{param}^\text{netw}$, for our purpose of encoding motion primitives  in NNs it is found that MLPs are still preferrable as function approximators over both SCNs~\cite{srouji2018structured} and also its extension FSCNs. MLPs are the focus in Experiment 4.

\subsection{Experiment 4: Neural network scheduling on velocity\label{subsec_expt4}}

\begin{table}
\centering
\caption{Experiment 4. Results when attempting to encode $N_\text{tasks}=14625$ training tasks \emph{all-at-once} (i.e., without network scheduling) for a MLP-[4,1,2], $T^\text{max}=1500$ and $(N_\text{restarts},N_\text{iter}^\text{max})=(5,5)$. $T_\text{learn}^\text{tot}=79504$~[s] translates to 22.1 hours of training time. 
\label{tab_ex4_ALLATONCE_soln}}
\def\arraystretch{0.8}
\begin{tabular}{|c|c|c|c|c|c|}
\hline 
MLP & $N_\text{tasks}^\star$ & $P^\star$ & $N_\text{param}^\text{netw}$ & $T_\text{learn}^\text{tot}$ & $N_\text{rest}^\star$\\
\hline\hline 
[4,1,2] & 13685 & -1209842 & 10 & 79504 & 0\\
\hline
\end{tabular}
\vspace{-0.4cm}
\end{table}

\begin{table}
\vspace{0.0cm}
\centering
\caption{Experiment 4. Results when encoding \emph{subsets} of training tasks that are scheduled on vehicle velocity $v_{x,0}$. For each subset $N_\text{tasks}=1125$ need to be encoded. The results for a 4D and 5D feature vector are compared. The scenarios for which not all training tasks could be solved are emphasized in bold. The percentage gain for \emph{final} $P^\star$ vs. the \emph{first} $P^\star$ solving all tasks is indicated by $\Delta P_\text{1st}^\star$ to stress benefits of refinement steps.\label{tab_ex4}}
\begin{align*}
\def\arraystretch{0.8}
& \begin{tabular}{|l|c|c|c|c|c|}
\hline
\multicolumn{6}{|c|}{MLP-[4,1,2]}\\\hline 
$v_{x,0}$ & $N_\text{tasks}^\star$ & $P^\star$ & $T_\text{learn}^\text{tot}$ & $N_\text{rest}^\star$ & $\Delta P_\text{1st}^\star$ \\
\hline\hline 
0  & 1125 & -4307  & 1283 & 5 & 1.7\%\\
10 & 1125 & -6818  & 2520 & 5 & 0.7\%\\
20 & 1125 & -19155  & 3478 & 5 & 11.7\%\\
30 & 1125 & -26804  & 2363 & 5 & 3.9\%\\
40 & 1125 & -36651  & 2221 & 5 & 17.4\%\\
50 & 1125 & -33751  & 2292 & 5 & 49.4\%\\
60 & 1125 & -47181  & 2362 & 2 & 23.2\%\\
70 & 1125 & -73455  & 2229 & 2 & 0.0\%\\
\textbf{80} & \textbf{1052} & \textbf{-151908} & \textbf{1879} & \textbf{0} & \textbf{--} \\
90 & 1125 & -141935  & 2046 & 1 & 0.0\%\\
\textbf{100} & \textbf{997} & \textbf{-224100} & \textbf{2230} & \textbf{0} & \textbf{--}\\
\textbf{110} & \textbf{595} & \textbf{-237900}  & \textbf{2244} & \textbf{0} & \textbf{--} \\
120 & 1125 & -151580  & 2069 & 1 & 0.0\%\\
\hline
\end{tabular}\\
\def\arraystretch{0.8}
& \begin{tabular}{|l|c|c|c|c|c|}
\hline
\multicolumn{6}{|c|}{MLP-[5,1,2]}\\\hline 
$v_{x,0}$ & $N_\text{tasks}^\star$ & $P^\star$ & $T_\text{learn}^\text{tot}$ & $N_\text{rest}^\star$ & $\Delta P_\text{1st}^\star$ \\
\hline\hline 
0   & 1125 & -3751  & 1263 & 5 & 6.7\%\\
10  & 1125 & -7454  & 3085 & 5 & 0.0\%\\
20  & 1125 & -13197  & 3392 & 5 & 11.3\%\\
30  & 1125 & -26805  & 2498 & 5 & 0.5\%\\
40  & 1125 & -43345  & 2157 & 5 & 1.5\% \\
50  & 1125 & -43961  & 2591 & 5 & 58.3\%\\
60  & 1125 & -52665  & 2347 & 1 & 11.5\%\\
70  & 1125 & -91471  & 2332 & 2 & 0.0\%\\
\textbf{80}  & \textbf{1016} & \textbf{-156711}  & \textbf{2016} & \textbf{0} & \textbf{--} \\
\textbf{90}  & \textbf{748} & \textbf{-191067}  & \textbf{2295} & \textbf{0} & \textbf{--} \\
100  & 1125 & -163466  & 2167 & 2 & 2.6\%\\
\textbf{110}  & \textbf{961} & \textbf{-260649}  & \textbf{2257} & \textbf{0} & \textbf{--} \\
120  & 1125 & -187651  & 1978 & 1 & 0.0\%\\
\hline
\end{tabular}
\end{align*}
\vspace{-0.35cm}
\end{table}

\begin{table}
\centering
\caption{Experiment 4. Illustration of how individual (more difficult) subsets of training tasks can conceptionally be reconsidered with different hyperparameters or, in general, even different network architectures in case that previously not all training tasks were solved. Here, for demonstration the hyperparameter setup is identical as for Table \ref{tab_ex4} and MLP-[4,1,2], except that $T^\text{max}=2000$ and $N_\text{iter}^\text{max}=10$. Note that results can be further fine-tuned individually. For example, a MLP-[4,1,2] trained for $T^\text{max}=1500$ and $(N_\text{restarts},N_\text{iter}^\text{max})=(10,10)$ for the subset of tasks scheduled on $v_{x,0}=110$ yields $P^\star=-183487$ (and $T_\text{learn}^\text{tot}=8673$ [s]).\label{tab_ex4_3missing}}
\def\arraystretch{0.8}
\begin{tabular}{|l|c|c|c|c|c|}
\hline 
$v_{x,0}$ & $N_\text{tasks}^\star$ & $P^\star$ & $T_\text{learn}^\text{tot}$ & $N_\text{rest}^\star$ & $\Delta P_\text{1st}^\star$ \\
\hline\hline 
80 & 1125 & -135610 & 5632 & 2/5 & 0.0\%\\
100 & 1125 & -199759 & 5120 & 5/5 & 12.9\%\\
110 & 1125 & -241227 & 5731 & 1/5 & 0.0\%\\
\hline
\end{tabular}
\vspace{-0.5cm}
\end{table}

\setlength\figureheight{6.5cm}
\setlength\figurewidth{6.5cm}
\begin{figure}
\centering
\includegraphics[scale=1.0]{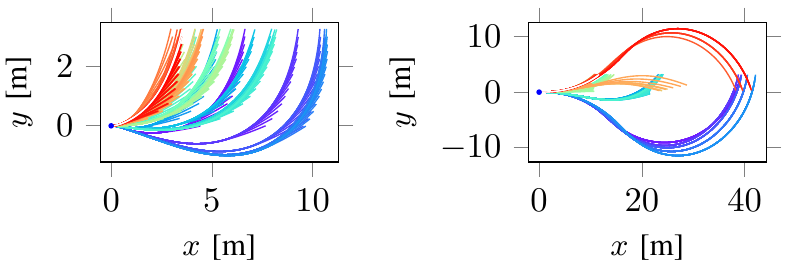}
\caption{Experiment 4. Display of all 375 learnt trajectories for a transition from $v_{x,0}=10$ to $v_{x}^\text{goal}=10$ (left) and from $v_{x,0}=50$ to $v_{x}^\text{goal}=40$ (right). Note that only for clarity 375 out of 1125 trajectories are displayed. For each subset of tasks $v_{x,0}$ is associated with $v_{x}^\text{goal}=v_{x,0}+\{-10,0,10\}$ (hence $3\cdot 375=1125$). See Sect. \ref{subsec_expt4} for further interpretations. }
\label{fig:ex4_traj}
\vspace{-0.5cm}
\end{figure}

Experiment 4 is characterized by (i) comparing feature vector $s_t$ from \eqref{eq_def_st_4D} and its 5D extension including $a_t[1]$, (ii) $T^\text{max}=1500$, (iii) a comparison of MLP-[4,1,2] and MLP-[5,1,2], (iv) $(N_\text{restarts},N_\text{iter}^\text{max})=(5,5)$, (v) attempting to solve a total of $N_\text{tasks}=14625$ training tasks at once and alternatively with network scheduling, whereby (vi) training tasks are generated by gridding $y^\text{goal}\in[0,3.5]$ and $v_{x,0}\in[0,120]$ with uniform spacings of 0.25m and 10km/h, respectively, $v_{x}^\text{goal}=v_{x,0}+\{-10,0,10\}$ for all $v_{x,0}$, $a_{-1}[0]\in\{-0.5, -0.25, 0, 0.25, 0.5\}$ and $a_{-1}[1]=a_{v}^\text{thres}+\{-0.4,-0.2,0,0.2,0.4\}$ (with $a_{v}^\text{thres}\approx 0.4$).

The results are summarized in Tables \ref{tab_ex4_ALLATONCE_soln}, \ref{tab_ex4} and \ref{tab_ex4_3missing} and Fig. \ref{fig:ex4_traj}. Several observations can be made. First, the benefits of network scheduling are illustrated. These include  (i) faster overall learning time (accumulated 8.1~hours for MLP-[4,1,2] in Table \ref{tab_ex4} vs. 22.1~hours in Table \ref{tab_ex4_ALLATONCE_soln}), and (ii) the ability to quicker detect difficult subsets of training tasks that can consequently also be resolved faster after modification of hyperparameters or even NN parametrization. Second, results of Table \ref{tab_ex4} suggest to prefer a 4D over a 5D $s_t$. The reduction of accumulated $P^\star$ for all 10 subsets of tasks solved completely by both MLP-[4,1,2] and MLP-[5,1,2] is 15\% for the former vs. the latter. Third, the capabilities of tiny NNs with only 10 parameters for MLP-[4,1,2] are demonstrated. These were tested both when (i) encoding all 14625 motion primitives at once, and (ii) learning subsets of 1125 tasks scheduled on vehicle velocity. It is remarkable that a MLP-[4,1,2] with 10 parameters (with $\theta_{VVC}$) can encode 13685 motion primitives to desired $\epsilon$-precision. Note that for the scheduling solution overall 13 MLPs are learnt, with \emph{each} having 10  parameters. Fourth, it is stressed that training tasks of Experiment 4 are not easy. Since these are generated uniformly, in the extreme case, the steering angle is initialized as $\delta_{0}=-20^\circ$ at an initial $v_{x,0}=120$km/h. This resulted in learnt trajectories with a maximum lateral overshoot of 62.1m before recovery of the desired $y^\text{goal}$. For visualization see also Fig. \ref{fig:ex4_traj} where the lateral maximum overshoot is slightly more than 10m for an initial $v_{x,0}=50$m/h. The left frame of Fig. \ref{fig:ex4_traj} is displayed to underline that for limited learning time only local optimal trajectories (according to the shortest path criterion subject to actuator and system constraints) are learnt. Trajectories are approximately balanced, but not entirely. Not displayed for brevity but very interestingly to observe was that for $v_{x,0}=0$ and $\delta_{0}<0^\circ$ trajectories were learnt that first \emph{reverse} drive for some time before only then accelerating forward. This fully makes sense given the initial negative tire heading, the pathlength minimization objective for $P^\star$ and $y^\text{goal}\geq 0$. Finally, as indicated in Table \ref{tab_ex4} and discussed above through the lateral overshoots, more difficult learning was observed for the higher velocity tasks. Here, tasks were setup (i) for demonstration that difficult motion primitives can still be encoded, and (ii) to motivate future work on automated extraction of meaningful motion primitives from real-world driving data, especially for higher vehicle velocities.

\section{Conclusion\label{sec_conclusion}}

Several methods were presented for efficient encoding of motion primitives in neural networks. Therefore in particular (i) specific virtual velocity constraints, (ii) neural network scheduling based on vehicle velocity, (iii) training task setups dismissing $[x^\text{goal},~\varphi^\text{goal}]$-related components and focusing on $y^\text{goal}$ for a 4D feature vector selection, and (iv) the capabilities of tiny neural networks were promoted. Furthermore, (i) a 3-states-2-controls kinematic and 16-states-2-controls dynamic model comparison, (ii) discussion of 3 feedforward neural network architectures including weighted skip connections, and (iii) implementation details of model-based and gradient-free training using 1 GPU were discussed. Findings were illustrated by means of 4 simulation experiments. 

Main subject of future work is closed-loop evaluation. This comprises robustness analysis for scenarios unseen during training, vehicle model mismatches, and analysis of the combination with a reference setpoint selector with obstacle avoidance inequality checks of closed-loop forward simulated trajectories in a receding horizon fashion. Here, the preferred 4D feature vector with focus on $y^\text{goal}$-selection is believed to enable the design of simple recursive waypoint selectors. Waypoints may also be concatenated to generate trajectory-trees that then permit very long planning horizons.

%

%
%
%
%
%
%

\nocite{*}
\bibliographystyle{ieeetr}
\bibliography{myref}

\appendix
\subsection{Dynamic 16-states vehicle model\label{subsec_appendix_dyn16vehmdl}}
The dynamic vehicle model is compactly summarized by \texttt{C++} code. System parameters are summarized in Table \ref{tab_hyperparam_16dynvehmdl}.

\begin{table}
\begin{center}
\vspace{0.3cm}
\caption{System parameters of the dynamic 16-states vehicle model.} 
\label{tab_hyperparam_16dynvehmdl}
\bgroup
\def\arraystretch{0.7}
\begin{tabular}{|l|l|l|l|}
\hline
Symbol &  Value & Unit & Comment\\ \hline \hline
$T_s$ & 0.01 & s & Sampling time\\
$\delta^\text{max}$ & 40 & deg & $\delta^\text{min}=-\delta^\text{max}$\\
$\dot{\delta}^\text{max}$ & 20 & deg/s & $\dot{\delta}^\text{min}=-\dot{\delta}^\text{max}$\\
$T_a^{\text{max}}$ & 1700 & Nm & Max. forward-drive torque \\
$T_a^{\text{min}}$ & -4000 &Nm & Braking and reverse-drive\\
$\dot{T}_a^{\text{max}}$ & 1700 & Nm/s & Rate forward-drive torque\\
$\dot{T}_a^{\text{min}}$ & -4000 & Nm/s & Rate brake and reverse-drive\\	
$m$ & 1450 & kg & Vehicle mass\\
$I_z$ & 2741.9 & kgm$^2$ & Moment of inertia\\
$I_w$ & 1.8 & kgm$^2$ & Moment of inertia\\	
$l_f$ & 1.1 & m & Dist. CoG to front-wheel axis\\
$l_r$ & 1.59 & m & Dist. CoG to rear-wheel axis\\
$h$   & 0.4  & m & Nominal vertical height CoG\\
$r_e$ & 0.3  & m & Effective tire radius\\
$g$  & 9.81  & m/s$^2$ & Gravitational acceleration\\	   
$k_s$ & 10000 & N/m & Spring constant \\
$c_s$ & 2000 &  Ns/m & Spring constant\\
$\frac{1}{2}\rho A_f c_d$ &  0.5*1.225*0.7 & N/m & Aerodynamic friction\\
$I_x$ & 500 & kgm$^2$ & Moment of inertia\\
$I_y$ & 2500 & kgm$^2$ & Moment of inertia\\
$l_w$ & 0.81 & m & Lateral dist. CoG to wheels\\
$B$ & 7.0 & - & Pacejka tire-model\\
$C$ & 1.6 & - & Pacejka tire-model\\
$D$ & 1.0 & - & Pacejka tire-model\\
\hline
\end{tabular}
\egroup
\end{center}
\vspace{-0.5cm}
\end{table}

\begin{lstlisting}
<@\textcolor{black}{//PRELIMINARY AUXILIARY ABBREVIATIONS.}@>
avthres=-1.0-2.0*Ta_min/(Ta_max-Ta_min);<@\textcolor{black}{//at this, Ta=0.}@>
Tsdotudeltamax=Ts*dotdelta_max/delta_max;
Tsdotuvmax=Ts*dotTa_max*2.0/(Ta_max-Ta_min);
Tsdotuvmin=Ts*dotTa_min*2.0/(Ta_max-Ta_min);
Tamaxmin05=0.5*(Ta_max-Ta_min); 
lf_div_lfplr=lf/(lf+lr); lr_div_lfplr=lr/(lf+lr);
mg05lr_div_lfplr=0.5*m*g*lr_div_lfplr;
mg05lf_div_lfplr=0.5*m*g*lf_div_lfplr;
<@\textcolor{red}{//SPECIAL CASE ($|v_x|<$1km/h and no ac/deceleration-command).}@>
if (z[3]<1/3.6 && z[3]>-1/3.6 && 
  a[1]<avthres+0.001 && a[1]>avthres-0.001){z[3]=0; 
  z[4]=0;z[5]=0;z[6]=0;z[7]=0;z[8]=0;z[9]=0; 
  z[10]=0;z[11]=0;z[12]=0;z[13]=0;z[14]=0;z[15]=0;    
<@\textcolor{red}{//REGULAR CASE (non-zero velocity, no singularity at zero velocity).}@>
}else{
<@\textcolor{blue}{//1. if zero-velocity then reinitialize $v_x$ of model to avoid pole-issues.}@>
if (z[3]<0.1/3.6 && z[3]>-0.1/3.6 && a[1]>avthres){ 
  z[3]=1/3.6;    z[10]=z[3]/Re; z[11]=z[3]/Re;
  z[12]=z[3]/Re; z[13]=z[3]/Re;
}else if (z[3]<0.1/3.6 && z[3]>-0.1/3.6 
&& a[1]<avthres){ z[3]=-1/3.6;   z[10]=z[3]/Re;
  z[11]=z[3]/Re;  z[12]=z[3]/Re; z[13]=z[3]/Re;}
<@\textcolor{blue}{//2. to deal with both forward/backward driving.}@>
sign_fb=1; if (z[3]<0){sign_fb=-1;} 
<@\textcolor{blue}{//3. controls: at this point both a[0] and a[1] in [-1,1].}@>
delta=delta_max*a[0];Ta=Ta_min+Tamaxmin05*(a[1]+1);
if (Ta>=0){ Ta1=0.5*Ta; Ta2=0.5*Ta;
  Tb1=0; Tb2=0; Tb3=0; Tb4=0;      
}else{ Ta1=0; Ta2=0; Tb1=-lf_div_lfplr*Ta; 
  Tb2=-lf_div_lfplr*Ta; Tb3=-lr_div_lfplr*Ta; 
  Tb4=-lr_div_lfplr*Ta;} 
<@\textcolor{blue}{//4. forces. }@>
beta=atan2f(z[4],z[3]);
<@\textcolor{blue}{//4.a. aerodynamic friction losses: $F_\text{air}$.}@>
Fair = rhoAfcd05*(z[3]*z[3] + z[4]*z[4]);
Fxair=Fair*cosf(beta);Fyair=Fair*sinf(beta);
<@\textcolor{blue}{//4.b. vertical forces: $F_{\eta 1},\dots,F_{\eta 4}$.}@>
Feta1=mg05lr_div_lfplr-ks*(z[14]-lf*sinf(z[8]) 
    +lw*sinf(z[6]))-cs*(z[15]-z[9]*lf*cosf(z[8]) 
    +lw*z[7]*cosf(z[6]));   
Feta2=mg05lr_div_lfplr-ks*(z[14]-lf*sinf(z[8])
    -lw*sinf(z[6]))-cs*(z[15]-z[9]*lf*cosf(z[8])
    -lw*z[7]*cosf(z[6])); 
Feta3=mg05lf_div_lfplr-ks*(z[14]+lf*sinf(z[8]) 
    +lw*sinf(z[6]))-cs*(z[15]+z[9]*lf*cosf(z[8])   
    +lw*z[7]*cosf(z[6])); 
Feta4=mg05lf_div_lfplr-ks*(z[14]+lf*sinf(z[8])
    -lw*sinf(z[6]))-cs*(z[15]+z[9]*lf*cosf(z[8])
    -lw*z[7]*cosf(z[6]));  
<@\textcolor{blue}{//4.c. tire-forces in the tire-aligned coordinate frame.}@>
<@\textcolor{blue}{//$\rightarrow F_{xw1},~F_{yw1}$.}@>
v1x=z[3]*cosf(beta-delta)/(cosf(beta))
    +z[5]*lf*sinf(delta)
    -z[5]*lw*cosf(beta-delta)/(cosf(beta));
s1x=(v1x-z[10]*Re)/v1x;
s1y=(z[3]*sinf(beta-delta)/(cosf(beta))
    +z[5]*lf*cosf(delta)
    +z[5]*lw*sinf(beta-delta)/(cosf(beta)))/v1x;
s1=sqrtf(s1x*s1x+s1y*s1y);
if (s1>0.001){<@\textcolor{black}{~//Bernoulli L'H{\^o}pital's rule to deal with singularities.}@>
  Fxw1=-sign_fb*s1x*D*sinf(C*atanf(B*s1))*Feta1/s1;
  Fyw1=-sign_fb*s1y*D*sinf(C*atanf(B*s1))*Feta1/s1;
}else{ Fxw1=0; Fyw1=0; }
<@\textcolor{blue}{//$\rightarrow F_{xw2},~F_{yw2}$.}@>
v2x=z[3]*cosf(beta-delta)/(cosf(beta))  
    +z[5]*lf*sinf(delta)
    +z[5]*lw*cosf(beta-delta)/(cosf(beta));
s2x=(v2x-z[11]*Re)/v2x;
s2y=(z[3]*sinf(beta-delta)/(cosf(beta))
    +z[5]*lf*cosf(delta)
    -z[5]*lw*sinf(beta-delta)/(cosf(beta)))/v2x;
s2=sqrtf(s2x*s2x + s2y*s2y);
if (s2>0.001){
  Fxw2=-sign_fb*s2x*D*sinf(C*atanf(B*s2))*Feta2/s2;
  Fyw2=-sign_fb*s2y*D*sinf(C*atanf(B*s2))*Feta2/s2;
}else{ Fxw2=0; Fyw2=0; }
<@\textcolor{blue}{//$\rightarrow F_{xw3},~F_{yw3}$.}@>
v3x=z[3]-z[5]*lw;       s3x=(v3x-z[12]*Re)/v3x; 
s3y=(z[4]-z[5]*lr)/v3x; s3=sqrtf(s3x*s3x+s3y*s3y);
if (s3>0.001){
  Fxw3=-sign_fb*s3x*D*sinf(C*atanf(B*s3))*Feta3/s3;
  Fyw3=-sign_fb*s3y*D*sinf(C*atanf(B*s3))*Feta3/s3;
}else{ Fxw3=0; Fyw3=0; }
<@\textcolor{blue}{//$\rightarrow F_{xw4},~F_{yw4}$.}@>
v4x=z[3]+z[5]*lw;       s4x=(v4x-z[13]*Re)/v4x;
s4y=(z[4]-z[5]*lr)/v4x; s4=sqrtf(s4x*s4x+s4y*s4y);
if (s4>0.001){
  Fxw4=-sign_fb*s4x*D*sinf(C*atanf(B*s4))*Feta4/s4;
  Fyw4=-sign_fb*s4y*D*sinf(C*atanf(B*s4))*Feta4/s4;
}else{ Fxw4 = 0; Fyw4 = 0; }
<@\textcolor{blue}{//4.d. forces in the vehicle-aligned frame.}@>
Fx1=(Fxw1*cosf(delta)-Fyw1*sinf(delta))*cosf(z[8])
    -Feta1*sinf(z[8]);
Fx2=(Fxw2*cosf(delta)-Fyw2*sinf(delta))*cosf(z[8]) 
    -Feta2*sinf(z[8]);
Fx3=Fxw3*cosf(z[8])-Feta3*sinf(z[8]);
Fx4=Fxw4*cosf(z[8])-Feta4*sinf(z[8]);
Fy1=(Fxw1*cosf(delta)
    -Fyw1*sinf(delta))*sinf(z[6])*sinf(z[8]) 
    +(Fyw1*cosf(delta)+Fxw1*sinf(delta))*cosf(z[6]) 
    +Feta1*sinf(z[6])*cosf(z[8]);
Fy2=(Fxw2*cosf(delta) 
    -Fyw2*sinf(delta))*sinf(z[6])*sinf(z[8]) 
    +(Fyw2*cosf(delta)+Fxw2*sinf(delta))*cosf(z[6])    
    +Feta2*sinf(z[6])*cosf(z[8]);
Fy3=Fxw3*sinf(z[6])*sinf(z[8])+Fyw3*cosf(z[6]) 
    +Feta3*sinf(z[6])*cosf(z[8]);
Fy4=Fxw4*sinf(z[6])*sinf(z[8])+Fyw4*cosf(z[6]) 
    +Feta4*sinf(z[6])*cosf(z[8]);     
<@\textcolor{blue}{//5. Euler forward: $z_{k+1}=z_k+T_sf(z_k,a_k,t_k)$.}@>
z[0]=z[0]+Ts*(z[3]*cosf(z[2])-z[4]*sinf(z[2])); 
z[1]=z[1]+Ts*(z[3]*sinf(z[2])+z[4]*cosf(z[2])); 
z[2]=z[2]+Ts*z[5]; 
z[3]=z[3]+Ts*((Fx1+Fx2+Fx3+Fx4-Fxair)/m+z[4]*z[5]);            
z[4]=z[4]+Ts*((Fy1+Fy2+Fy3+Fy4-Fyair)/m-z[3]*z[5]);             
z[5]=z[5]+Ts*(lf*(Fy1+Fy2)-lr*(Fy3+Fy4)
     +lw*(Fx2+Fx4-Fx1-Fx3))/Iz; 
z[6]=z[6]+Ts*z[7];                                                         
z[7]=z[7]+Ts*( lw*(Feta1+Feta3-Feta2-Feta4)
     +h*(Fy1+Fy2+Fy3+Fy4) )/Ix; 
z[8]=z[8]+Ts*z[9];                                                  
z[9]=z[9]+Ts*( lr*(Feta3+Feta4)-lf*(Feta1+Feta2) 
     -h*(Fx1+Fx2+Fx3+Fx4) )/Iy;
z[10]=z[10]+Ts*(Ta1-Tb1-Re*Fxw1)/Iw;                                     
z[11]=z[11]+Ts*(Ta2-Tb2-Re*Fxw2)/Iw;                              
z[12]=z[12]+Ts*(0-Tb3-Re*Fxw3)/Iw;                                    
z[13]=z[13]+Ts*(0-Tb4-Re*Fxw4)/Iw;                                    
z[14]=z[14]+Ts*z[15];
z[15]=z[15]+Ts*((Feta1+Feta2+Feta3+Feta4)/m - g);                                    
<@\textcolor{blue}{//6. ensure angles in $[0, 2\pi]$.}@>
while (z[2]>2*pi){z[2]=z[2]-2*pi;}
while (z[2]<0){   z[2]=z[2]+2*pi;}
while (z[6]>2*pi){z[6]=z[6]-2*pi;}
while (z[6]<0){   z[6]=z[6]+2*pi;}
while (z[8]>2*pi){z[8]=z[8]-2*pi;}
while (z[8]<0){   z[8]=z[8]+2*pi;}}
\end{lstlisting}


\end{document}